\documentclass[11pt]{article}
\usepackage{algorithmic,amsfonts,amsmath,amsthm,xspace}

\makeatletter
\newfont{\footsc}{cmcsc10 at 8truept}
\newfont{\footbf}{cmbx10 at 8truept}
\newfont{\footrm}{cmr10 at 10truept}

\makeatother 

\newtheorem{observation}{Observation}
\newtheorem{corollary}{Corollary}
\newtheorem{theorem}{Theorem}

\newtheorem{lemma}{Lemma}
\newcommand{\ignore}[1]{}
\def\eps{\epsilon}
\def\B{\ensuremath{\mathbb{B}}}
\def\S{\ensuremath{\mathbb{S}}}

\def\del{\nabla}

\def\vol{\operatorname{vol}}

\def\reals{\ensuremath{\mathbb{R}}}

\def\cc{\colon\thinspace}
\def\st{\thinspace\big|\thinspace}
\def\given{\thinspace\big|\thinspace}

\def\E{\operatorname{\bf E}}
\def\P{\operatorname{\bf P}}

\ignore{11pt, 12 pages for long versions, 3 pages for short incl.
bibliography.  Instructions for submissions are available at
http://www.sigact.acm.org/~soda05/SODA05.html . The deadline for
submission is 5:00 PM EDT, July 5, 2004.

Note:  appendix may count against page limit this year, due to
authors "abusing the appendix rules" in the past.  --Abie}

\begin{document}

\title{Online convex optimization in the bandit setting:\\ gradient descent
without a gradient}

\author{Abraham D.~Flaxman \and Adam
Tauman Kalai \and H.~Brendan McMahan}
\maketitle              

\begin{abstract}
We study a general online convex optimization problem.  We have a
convex set $S$ and an unknown sequence of cost functions
$c_1,c_2,\ldots,$ and in each round, we choose a feasible point
$x_t$ in $S$, and learn the cost $c_t(x_t)$.  If the function
$c_t$ is also revealed after each round then, as Zinkevich shows
in \cite{Z}, gradient descent can be used on these functions to
get regret bounds of $O(\sqrt{n})$. That is, after $n$ rounds, the
total cost incurred will be $O(\sqrt{n})$ more than the cost of
the best single feasible decision chosen with the benefit of
hindsight, $\min_{x\in S} \sum c_t(x)$.

We extend this to the ``bandit'' setting where each period, only
the cost $c_t(x_t)$ is revealed, and bound the expected regret
(against an oblivious adversary) as $O(n^{5/6})$.

Our approach uses a simple approximation of the gradient that is
computed from evaluating $c_t$ at a single (random) point.  We
show that this biased estimate is sufficient to approximate
gradient descent on the sequence of functions.  In other words, it
is possible to use gradient descent in the online setting without
seeing anything more than the value of the functions at a single
point.

For the online linear optimization problem \cite{KV}, algorithms
with low regrets in the bandit setting have recently been given
against oblivious \cite{AK} and adaptive adversaries \cite{MB}. In
contrast to these algorithms, which divide time into explicit {\em
explore} and {\em exploit} phases, our algorithm can be
interpreted as doing a small amount of exploration in each round.
\end{abstract}

\section{Introduction}
Consider three optimization settings where one would like to minimize a
convex function (equivalently maximize a concave function).  In
all three settings, gradient descent is one of the most popular
methods.
\begin{enumerate}
\item Offline: Minimize a fixed convex cost function $c\cc\reals^d
\rightarrow \reals$.  In this case, gradient descent is
$x_{t+1}=x_t-\eta \del c(x_t)$.

\item Stochastic: Minimize a fixed convex cost function $c$ given
only ``noisy'' access to $c$, for example, we can only get
$c_t(x)=c(x)+\eps_t(x)$ for zero-mean error random error
$\eps_t(x)$. Here, stochastic gradient descent is
$x_{t+1}=x_t-\eta \del c_{t}(x_t)$. (The intuition is that the
expected gradient is correct, i.e.~$\E[\del c_t(x)]=\del
\E[c_t(x)]=\del c(x)$.)  In non-convex cases, the additional
randomness may actually help avoid local minima \cite{BT}, in a
manner similar to  Simulated Annealing \cite{K+}.

\item Online: Minimize an unknown sequence of convex functions,
$c_1,c_2,\ldots,$ i.e.~choose a sequence $x_1,x_2,\ldots$ where
each $x_t$ only depends on $x_1,x_2,\ldots, x_{t-1}$ and
$c_1,c_2,\ldots,c_{t-1}$.  The goals is to have low {\em regret}
$\sum c_t(x_t)- \min \sum c_t(x)$ for not using the best single
point, chosen with the benefit of hindsight.  In this setting,
Zinkevich analyzes the regret of gradient descent given by
$x_{t+1}=x_t -\eta \del c_t(x_t)$.
\end{enumerate}

We will focus primarily on gradient descent in a ``bandit''
version of the online setting.  As a motivating example, consider
a company that has to decide, every week, how much to spend
advertising on each of a $d$ different channels, represented as a
vector $x_t \in \reals^d$. At the end of each week, they calculate
their total profit $p_t(x_t)$. In the offline case, one might
assume that each week the function $p_1, p_2, \ldots$ are
identical. In the stochastic case, one might assume that different
weeks will have profit functions, but the $p_t(x)$ will be noisy
realizations of some true underlying profit function, for example
$p_t(x) = p(x) + \eps_t(x)$, where $\eps_t(x)$ has mean 0. In the
online case, \emph{no assumptions} are made about a distribution
over convex profit functions and instead they are modeled as the
malicious choices of an (oblivious) adversary.  This allows, for
example, for the possibility of a bad economy which cause the
profits to crash.

In this paper, we consider the bandit case where we only have
black-box access to the function(s) and thus cannot access the
gradient of $c_t$ directly for gradient descent.  (In the
advertising example, the advertisers only find out the total
profit of their chosen $x_t$, and not how much they would have
profited from other values of $x$.) This type of optimization is
sometimes referred to as {\em direct} or {\em gradient-free}.

A natural approach in the black-box case, for all three settings,
would be to estimate the gradient by evaluating the function at
several places around the point, and from them estimate the
gradient (see Finite Difference Stochastic Approximation,
e.g.~Chapter 6 of \cite{S2}).  However, in the online setting, the
functions change adversarially over time and we only can evaluate
each function once.  We use a one-point estimate of the gradient
to sidestep these difficulties.

\subsection{A one-point estimate to the gradient}
Our estimate is based on the observation that for a uniformly
random unit vector $u$,
\begin{eqnarray}
\del f(x) &\approx& \E\left[\bigl(f(x+\delta
u)-f(x)\bigr)u\right]d/\delta\label{iq0}\\
&=& \E[f(x+\delta u)u]d/\delta\label{iq1}
\end{eqnarray}
The first line looks more like an approximation of the gradient
than the second. But because $u$ is uniformly random over the
sphere, in expectation the second term in the first line is zero.
Thus, it would seem that {\em on average}, the vector
$(d/\delta)f(x+\delta u)u$ is an estimate of the gradient with low
bias, and thus we say loosely that it is an approximation to the
gradient.

To make this precise,  we show in Section \ref{sec:approx} that
$(d/\delta)f(x+\delta u)u$ is an unbiased estimator the gradient of a {\em
smoothed} version of $f$, where the value of at $x$ is replaced by
the average over a ball of radius $\delta$ around $x$.  For a
vector $v$ selected uniformly at random from the unit ball, let
    $$
    \hat{f}(x) = \E[f(x+\delta v)].
    $$
Then
    $$
    \del \hat{f}(x)  = \E[f(x+\delta u)u]d/\delta.
    $$
Interestingly, this does not require that $f$ be differentiable.

Our method of obtaining  a one-point estimate of the gradient is
similar to a one-point estimates proposed independently by by
Granichin \cite{G} and Spall \cite{S}.  Spall's estimate uses a
perturbation vector $p$, in which each entry is a zero-mean
independent random variable, to produce an estimate of the
gradient
    $\hat{g}(x) = \frac{f(x+\delta
    p)}{\delta}\left[\frac{1}{p_1},\frac{1}{p_2},\ldots,\frac{1}{p_d}\right]^T.$
This estimate is more of a direct attempt to estimate the gradient
coordinatewise and is not rotationally invariant. Spall's analysis
focuses on the stochastic setting and requires that the function
is three-times differentiable. In \cite{G2}, Granichin shows that
a similar approximation is sufficient to perform gradient descent
in a very general stochastic model.
    \ignore{
    I took out the comment about two-point estimates because they are a whole area
    in themselves.  If we want to mention them, we shouldn't credit
    them to Granichin.  --Adam

    OK, let's not mention them, then.  --Abie
    }

Unlike \cite{G,G2,S}, we work in an adversarial model, where
instead of trying to make the restrictions on the randomness of
nature as weak as possible, we pessimistically assume that nature
is conspiring against us.  Even in the (oblivious) adversarial setting a
one-point estimate of the gradient is sufficient to make gradient
descent work.

\subsection{Guarantees and analysis outline}
We use the following online bandit version of Zinkevich's model.
There is a fixed unknown sequence of convex functions $c_1, c_2,
\ldots, c_n\cc S \rightarrow [- C,C]$, where $C>0$ and $S
\subseteq \reals^d$ is a convex feasible set.  The decision-maker
sequentially chooses points $x_1,x_2,\ldots,x_n \in S$. After
$x_t$ is chosen, the value $c_t(x_t)$ is revealed, and $x_{t+1}$
must be chosen only based on $x_1,x_2,\ldots,x_t$ and
$c_1(x_1),c_2(x_2),\ldots,c_t(x_t)$ (and private randomness).

Zinkevich shows that, when the gradient $\del c_t(x_t)$ is given
to the decision-maker after each round, an online gradient descent
algorithm guarantees,
\begin{equation}
\mbox{regret}=\sum_{t=1}^n c_t(x_t)- \min_{x\in S} \sum_{t=1}^n
c_t(x)
     \leq DG\sqrt{n}.\label{marty}
\end{equation}
Here $D$ is the diameter of the feasible set, and $G$ is an upper
bound on the magnitudes of the gradients.

By elaborating on his technique, we present update rules for
computing a sequence of $x_{t+1}$ in the absence of $\del
c_t(x_t)$, that give the following guarantee on expected regret:
 $$
     \E\bigg[\sum_{t=1}^n c_t(x_t)\bigg]
     -\min_{x \in S} \sum_{t=1}^n c_t(x)
     \leq 6n^{5/6}dC$$
Notice we have replaced the differentiability and bounded gradient
assumptions by bounded function assumptions.  As expected, our
guarantees in the bandit setting are worse than those of the
full-information setting: $O(n^{5/6})$ instead of $O(n^{1/2})$. If
we make an additional assumption that the functions satisfy an
$L$-Lipschitz condition (which is less restrictive than a bounded
gradient assumption), then we can reduce expected regret to
$O(n^{3/4})$:
$$\E\bigg[\sum_{t=1}^n c_t(x_t)\bigg]
     -\min_{x \in S} \sum_{t=1}^n c_t(x)
    \leq 6n^{3/4}d\left(\sqrt{CLD}+C\right).
    $$
To prove these bounds, we have several pieces to put together.
First of all, we show that Zinkevich's guarantee (\ref{marty})
holds unmodified for vectors that are unbiased estimates of the
gradients.  Here $G$ becomes an upper bound on the magnitude of
the estimates.

Now, the updates should roughly be of the form $x_{t+1}=x_t-\eta
(d/ \delta)\E[c_t(x_t+\delta u_t)u_t]$.  Since we can only
evaluate each function at one point, that point should be
$x_t+\delta u_t$. However, our analysis applies to bound $\sum
c_t(x_t)$ and not $\sum c_t(x_t+\delta u_t)$.  Fortunately, these
points are close together and thus these values should not be too
different.

Another problem that arises is that the perturbations may move
points outside the feasible set.  To deal with these issues, we
stay on a subset of the set such that the ball of radius $\delta$
around each point in the subset is contained in $S$.  In order to
do this, it is helpful to have bounds on the radii $r,R$ of balls
that are contained in $S$ and that contain $S$, respectively.
Then guarantees can be given in terms of $R/r$.  Finally, we can
use existing algorithms \cite{LV} to reshape the body so $R/r \leq
d$ to get the final results.

\subsection{Related work}
For direct offline optimization, i.e.~from an oracle that
evaluates the function, in theory one can use the ellipsoid
\cite{ellipsoid} or more recent random-walk based approaches
\cite{BV}.  In black-box optimization, practitioners often use
Simulated Annealing \cite{K+} or finite difference/simulated
perturbation stochastic approximation methods (see, for example,
\cite{S2}). In the case that the functions may change dramatically
over time, a single-point approximation to the gradient may be
necessary. Granichin and Spall propose a different single-point
estimate of the gradient \cite{G,S}.

In addition to the appeal of an online model of convex
optimization, Zinkevich's gradient descent analysis can be applied
to several other online problems for which gradient descent and
other special-purpose algorithms have been carefully analyzed,
such as Universal Portfolios \cite{C,HSSW,KV00}, online linear
regression \cite{KW}, and online shortest paths \cite{TW02} (one
convexifies to get an online shortest flow problem).

A similar line of research has developed for the problem of online
linear optimization \cite{KV,AK,MB}.  Here, one wants to solve the
related but incomparable problem of optimizing a sequence of
linear functions, over a possibly non-convex feasible set,
modeling problems such as online shortest paths and online binary
search trees (which are difficult to convexify).  Kalai and
Vempala \cite{KV} show that, for such linear optimization problems
in general, if the offline optimization problem is solvable
efficiently, then regret can be bounded by $O(\sqrt{n})$ also by
an efficient online algorithm, in the full-information model.
Awerbuch and Kleinberg \cite{AK} generalize this to the bandit
setting against an oblivious adversary (like ours).  Blum and
McMahan \cite{MB} give a simpler algorithm that applies to {\em
adaptive} adversaries, that may choose their functions $c_t$
depending on the previous points.

A few comparisons are interesting to make with the online linear
optimization problem.  First of all, for the bandit versions of
the linear problems, there was a distinction between exploration
phases and exploitation phases.  During exploration phases, one
action from a {\em barycentric spanner} \cite{AK} basis of $d$
actions was chosen, for the sole purpose of estimating the linear
objective function.  In contrast, our algorithm does a little bit
of exploration each time. Secondly, Blum and McMahan \cite{MB}
were able to compete against an adaptive adversary, using a
careful Martingale analysis.  It is not clear if that can be done
in our setting.

\subsection{Notation} Let $\B$ and $\S$ be the unit ball and sphere
centered around the origin in $d$ dimensions, respectively,
     \begin{align*}
     \B &= \{x \in \reals^d \st |x|\leq 1\}\\
     \S &= \{x \in \reals^d \st |x|=1\}
     \end{align*}
The ball and sphere of radius $a$ are $a \B$ and $a \S$,
correspondingly.

The sequence of functions $c_1,c_2, \ldots c_n\cc S \rightarrow
\reals$ are fixed in advance (we only handle such an oblivious
adversary, not an adaptive one). The sequence of points we pick is
$x_1,x_2, \ldots,x_n$.  For bandit algorithms, we need to be
randomized, so we consider our {\em expected regret}:
$$\E\bigg[\sum_{t=1}^n c_t(x_t)\bigg]-\min_z
\sum_{t=1}^n c_t(z).$$

Zinkevich assumes the existence of a projection oracle $\P_S(x)$, projecting the point $x$ onto the nearest
point in the convex set $S$,
$$\P_S(x)=\arg\min_{z \in S} |x-z|.$$
Projecting onto the set is an elegant way to handle the situation that the gradient takes one outside of the set, and is a common trick in the optimization literature.  Note that computing $\P_S$ is ``only'' an offline convex
optimization problem.  While for arbitrary feasible sets, this may
seem difficult, for standard shapes, such as cube, ball, simplex,
etc., the calculation is quite straightforward.

A function $f$ is $L$-Lipschitz if
$$|f(x)-f(y)| \leq L|x-y|,$$
for all $x,y$ in the domain of $f$.

   We assume $S$ contains the ball of radius $r$
centered at the origin and is contained in the ball of radius $R$,
i.e.,
$$ r \B \subseteq S \subseteq R\B.$$

\section{Approximating the gradient with a single
sample}\label{sec:approx}

The main observation of this section is that we can estimate the
gradient of a function $f$ by taking a random unit vector $u$ and
scaling it by $f(x+\delta u)$, i.e.~$\hat{g} = f(x+\delta u)u$.
The approximation is correct in the sense that $\E[\hat{g}]$ is
proportional to the gradient of a {\em smoothed} version of $f$.
For any function $f$, for $v$ random from the unit ball, define
\begin{equation}\hat{f}(x) = \E_{v \in \B}[f(x+\delta v)].\label{eq:hbm}\end{equation}

\begin{lemma}\label{lem1}
Fix $\delta>0$, over random unit vectors $u$,
     $$
     \E_{u \in \S}[f(x+\delta u)u] = \frac{\delta}{d}\del \hat{f}(x).
     $$
\end{lemma}

\begin{proof}
If $d=1$, then the fundamental theorem of calculus implies,
$$\frac{d}{dx}\int_{-\delta}^{\delta} f(x+v)dv = f(x+\delta)-f(x-\delta).$$
The $d$-dimensional generalization, following from Stoke's
theorem, is,
\begin{equation}
\del \int_{\delta \B} f(x+v)dv = \int_{\delta \S}
f(x+u)\frac{u}{\|u\|} du.\label{sto}
\end{equation}
By definition,
     \begin{equation}\label{eq:11}
     \hat{f}(x) = \E[f(x+\delta v)] = \frac{\int_{\delta\B}
f(x+v)dv}{\vol_d(\delta
     \B)}.
     \end{equation}
Similarly,
     \begin{equation}\label{eq:22}
     \E[f(x+\delta u)u] = \frac{\int_{\delta\S} f(x+u)\cdot \frac{u}{\|u\|}
du}{\vol_{d-1}(\delta
     \S)}.
     \end{equation}
Combining Eq.'s (\ref{sto}), (\ref{eq:11}), and (\ref{eq:22}), and
the fact that ratio of volume to surface area of a $d$-dimensional
ball of radius $\delta$ is $\delta/d$ gives the lemma.
\end{proof}

Notice that the function $\hat{f}$ is differentiable even when $f$
is not.

\section{Expected Gradient Descent}\label{sec:egd}

First we consider a version of gradient descent where each step
$t$ we get a random vector $g_t$ with expectation equal to the
gradient.  Then we can still use Zinkevich's online analysis of
gradient descent.  For lack of a better choice, we use the
starting point $x_1=0$, the center of a containing ball of radius
$R\leq D$ and $x_{t+1}=\P_S(x_t-\eta g_t)$.
\begin{lemma}\label{orange}
Let $c_1,c_2,\ldots,c_n\cc S \rightarrow \reals$ be a sequence of
convex, differentiable functions.  Let $x_1,x_2,\ldots,x_n \in S$ be defined by
$x_1=0$ and $x_{t+1}=\P_S(x_t-\eta g_t)$, where $\eta
 >0$ and $g_1,\dots,g_n$ are vector-valued random variables with
$\E[g_t \given x_t] = \del c_t(x_t)$ and $\|g_t\|\leq G$, for some $G>0$
(this also implies $\|\del c_t(x)\|\leq G$).  Then,
for $\eta = \frac{R}{G\sqrt{n}}$,
     $$
     \E\bigg[\sum_{t=1}^n c_t(x_t)\bigg]- \min_{x\in S} \sum_{t=1}^n c_t(x)
     \leq RG\sqrt{n}.
     $$
\end{lemma}

\begin{proof}
Let $x_\star$ be a point in $S$ minimizing $\sum_{t=1}^n c_t(x)$.

Since $c_t$ is convex and differentiable, we can bound the
difference between $c_t(x_t)$ and $c_t(x_\star)$ in terms of the
gradient.
     \begin{align*}
     c_t(x_t)-c_t(x_\star) &\leq \del c_t(x_t) \cdot (x_t-x_\star)\\
     &= \E[g_t\given x_t] \cdot (x_t-x_\star)\\
     &= \E[g_t \cdot (x_t-x_\star) \given x_t]
     \end{align*}
Taking the expectation on both sides of this inequality yields
     \begin{equation}
     \E[c_t(x_t)-c_t(x_\star)] \leq \E[g_t \cdot (x_t-x_\star)].
     \label{eq:1}
     \end{equation}
Following Zinkevich's analysis, we use $\|x_t-x_\star\|^2$ as a
potential function.  Since $S$ is convex, for any $x \in \reals^d$
we have $\|\P_S(x)-x_\star\| \leq \|x-x_\star\|$.  So
     \begin{align*}
     \|x_{t+1}-x_\star\|^2
     &= \|\P_S(x_t-\eta g_t) - x_\star\|^2\\
     &\leq \| x_t - \eta g_t - x_\star\|^2\\
     &= \|x_t-x_\star\|^2 + \eta^2 \|g_t\|^2 - 2 \eta (x_t-x_*) \cdot g_t\\
     &\leq \|x_t-x_\star\|^2 + \eta^2 G^2 - 2 \eta (x_t-x_\star) \cdot
     g_t.
     \end{align*}
After rearranging terms, we have
     \begin{equation}\label{eq:2}
     g_t \cdot (x_t-x_\star)
     \leq \frac{\|x_t-x_\star\|^2 - \|x_{t+1}-x_\star\|^2 + \eta^2G^2}{2\eta}.
     \end{equation}
By putting Eq. (\ref{eq:1}) and Eq. (\ref{eq:2}) together we see
that
     \begin{align*}
     \E\bigg[ \sum_{t=1}^n c_t(x_t)\bigg] - \sum_{t=1}^n c_t(x_\star)
     &= \sum_{t=1}^n \E[ c_t(x_t) - c_t(x_\star)]\\
     &\leq \sum_{t=1}^n \E[g_t\cdot (x_t-x_\star)]\\
     &\leq \sum_{t=1}^n \E\left[ \frac{\|x_t-x_\star\|^2 -
\|x_{t+1}+x_\star\|^2 + \eta^2G^2}{2\eta} \right]\\
     &= \E\left[\frac{\|x_1-x_\star\|^2}{2\eta} +
n\frac{\eta^2G^2}{2\eta}\right]\\
     &\leq \frac{R^2}{2\eta}+n\frac{\eta G^2}{2}.
     \end{align*}
The last step follows because we chose $x_1=0$ and $S \subseteq
R\B$.  Plugging in $\eta=R/G\sqrt{n}$ gives the lemma.
\end{proof}

\subsection{Algorithm and analysis}
In this section, we analyze the algorithm given in Figure
\ref{fig:alg}.
     \begin{figure}[h]
     \hrule

     \medskip

     $\mathrm{BGD}(\alpha,\delta,\nu)$
 \begin{itemize} \itemsep 1pt
 \item $y_1=0$
 \item At each period $t$:
 \begin{itemize} \itemsep 1pt
    \ignore{
    Abie -- I couldn't get the algorithmic thing to work. Also, I didn't like the for loop, since the algorithm
    itself doesn't presuppose a fixed horizon $n$.  --Adam

    OK.  I think the itemize is ugly, but it's not a big deal
    either way.  --Abie
    }
 \item select unit vector $u_t$ uniformly at random
 \item $x_t := y_t+\delta u_t$
 \item $y_{t+1} := \P_{(1-\alpha)S}(y_t-\nu c_t(x_t)u_t)$
 \end{itemize}
  \end{itemize}
    \hrule
     \caption{Bandit gradient descent algorithm}\label{fig:alg}
     \end{figure}

We begin with a few observations.
\begin{observation}\label{O1}
The optimum in $(1-\alpha)S$ is near the optimum in $S$,
$$\min_{x \in (1-\alpha)S} \sum_{t=1}^n c_t(x) \leq 2 \alpha C n + \min_{x \in
S}\sum_{t=1}^n c_t(x).$$
\end{observation}
\begin{proof}
Clearly $(1-\alpha)S \subseteq S$.  Also,
     $$
     \min_{x \in (1-\alpha)S} \sum_{t=1}^n c_t(x)
     = \min_{x \in S}\sum_{t=1}^n c_t\bigl((1-\alpha)x\bigr).
     $$
And since each $c_t$ is convex and $0\in S$, we have
     \begin{align*}
     \min_{x \in S}\sum_{t=1}^n c_t\bigl((1-\alpha)x\bigr)
     &\leq \min_{x \in S} \sum_{t=1}^n \alpha c_t(0)+(1-\alpha)c_t(x)\\
     &= \min_{x \in S} \sum_{t=1}^n \alpha(c_t(0)-c_t(x))+c_t(x).
     \end{align*}
Finally, since for any $y\in S$ and $t\in \{1,\dots,n\}$ we have
$|c_t(y)| \leq C$, we may conclude that
     $$
     \min_{x \in S} \sum_{t=1}^n \alpha(c_t(0)-c_t(x))+c_t(x)
     \leq \min_{x \in S} \sum_{t=1}^n \alpha 2C+c_t(x).
     $$
\end{proof}

\begin{observation}\label{O2}
For any point $x$ in $(1-\alpha)S$ the ball of radius $\alpha r$
centered at $x$ is contained in $S$.
\end{observation}
\begin{proof}
Since $r\B \subseteq S$ and $S$ is convex, we have
$$(1-\alpha)S+\alpha r \B \subseteq (1-\alpha)S+\alpha S = S.$$
\end{proof}

The next observation establishes a bound on the maximum the
function can change in $(1-\alpha)S$, an effective Lipschitz
condition.
\begin{observation}\label{O3}
For any $x$ in $(1-\alpha)S$ and any  $y$ in $S$
     $$
     |c_t(x)-c_t(y)|\leq \frac{2C}{\alpha r}|x-y|.
     $$
\end{observation}
\begin{proof}
Let $y = x+\Delta$.  If $|\Delta|>\alpha r$, the observation
follows from $|c_t|<C$.  Otherwise, let $z=x+\alpha r
\frac{\Delta}{|\Delta|}$, the point at distance $\alpha r$ from
$x$ in the direction $\Delta$.  By the previous observation, we
know $z \in S$.  Also, $y = \frac{|\Delta|}{\alpha r} z +
\left(1-\frac{|\Delta|}{\alpha r}\right)x$, so,
     \begin{align*}
     c_t(y)
     &\leq \frac{|\Delta|}{\alpha r}c_t(z)+\left(1-\frac{|\Delta|}{\alpha
r}\right)c_t(x)\\
     &= c_t(x) + \frac{c_t(z)-c_t(x)}{\alpha r}|\Delta|\\
     &\leq c_t(x) + \frac{2C}{\alpha r} |\Delta|. \qedhere
     \end{align*}
\end{proof}

Now we are ready to select the parameters.
\begin{theorem}\label{thm:param}
For any $n\geq\left(\frac{3Rd}{2r}\right)^2$ and $\nu
=\frac{R}{C\sqrt{n}}$, $\delta=\sqrt[3]{\frac{rR^2d^2}{12n}}$, and
$\alpha=\sqrt[3]{\frac{3Rd}{2r\sqrt{n}}}$, the expected regret of
$\mathrm{BGD}(\nu,\delta,\alpha)$ is upper bounded by
     $$
     \E\bigg[\sum_{t=1}^n c_t(x_t)\bigg] - \min_{x \in S} \sum_{t=1}^n
c_t(x) \leq 3
     C n^{5/6}\sqrt[3]{dR/r}.
     $$
\end{theorem}
\begin{proof}
We begin by showing that the points $x_t \in S$.  Since $y_t \in
(1-\alpha)S$, Observation \ref{O2} implies this fact as long as
$\frac{\delta}{r} \leq \alpha < 1$, which is the case for
$n\geq(3Rd/2r)^2$.

Suppose we wanted to run the gradient descent algorithm on the
functions $\hat{c}_t$ defined by (\ref{eq:hbm}), and the set $(1-\alpha)S$.
If we let
     $$
     g_t = \frac{d}{\delta} c_t(x_t+\delta u_t)u_t
     $$
then  (since $u_t$ is selected uniformly at random from $\S$)
Lemma \ref{lem1} says $\E[g_t \given x_t] = \del\hat{c}_t(x_t)$.
So Lemma \ref{orange} applies with the update rule:
     $$
     x_{t+1}
     = \P_{(1-\alpha)S}(x_t - \eta g_t)
     = \P_{(1-\alpha)S}(x_t - \eta\frac{d}{\delta} c_t(x+\delta
     u_t)u_t),
     $$
which is exactly the update rule we are using to obtain $y_t$,
with $\eta = \nu \delta/d$.  Since
     $$
     \|g_t\|
     = \left\|\frac{d}{\delta} c_t(x+\delta u_t)u_t\right\|
     \leq dC/\delta,
     $$
we can apply Lemma \ref{orange} with $G=dC/\delta$.  By our choice
of $\nu$, we have $\eta = R/G\sqrt{n}$, and so the expected regret
is upper bounded by
     $$
     \E\bigg[\sum_{t=1}^n \hat{c}_t(y_t)\bigg]
     -\min_{x\in (1-\alpha)S} \sum_{t=1}^n \hat{c}_t(x)
     \leq \frac{RdC\sqrt{n}}{\delta}.
     $$

Let $L=\frac{2C}{\alpha r}$, which will act an ``effective
Lipschitz constant''.  Notice that for $x \in (1-\alpha)S$
Observation \ref{O3} shows that $|\hat{c}_t(x)-c_t(x)| \leq \delta
L$ since $\hat{c}_t$ is an average over inputs within $\delta$ of
$x$.  Since $|y_t-x_t|=\delta$, Observation \ref{O3} also shows
that
     $$
     |\hat{c}_t(y_t)-c_t(x_t)|
     \leq |\hat{c}_t(y_t)-c_t(y_t)| + |c_t(y_t)-c_t(x_t)|
     \leq 2 \delta L.
     $$
These with the above imply,
     $$
     \E\bigg[\sum_{t=1}^n \bigl(c_t(x_t)-2\delta L\bigr)\bigg]
     - \min_{x \in (1-\alpha)S} \sum_{t=1}^n\bigl(c_t(x)+\delta L\bigr)
     \leq
     \frac{RdC\sqrt{n}}{\delta},
     $$
so rearranging terms and using Observation \ref{O1} gives
     \begin{equation}
     \E\bigg[\sum_{t=1}^n c_t(x_t)\bigg]
     - \min_{x \in S} \sum_{t=1}^n c_t(x)
     \leq
     \frac{RdC\sqrt{n}}{\delta}+3\delta L n+ 2\alpha Cn.
\label{brendanwillneverreadthis}
\end{equation}
Plugging in $L=\frac{2C}{\alpha r}$ gives,
     $$
     \E\bigg[\sum_{t=1}^n c_t(x_t)\bigg]
     -\min_{x \in S} \sum_{t=1}^n c_t(x)
     \leq
     \frac{RdC\sqrt{n}}{\delta}+\frac{\delta}{\alpha} \frac{6Cn}{r}+ \alpha
     2Cn.
     $$
This expression is of the form $\frac{a}{\delta}+b
\frac{\delta}{\alpha} + c\alpha$.  Setting $\delta =
\sqrt[3]{\frac{a^2}{bc}}$ and $\alpha = \sqrt[3]{\frac{ab}{c^2}}$
gives a value of $3\sqrt[3]{abc}$.  The lemma is achieved for $a
=RdC\sqrt{n}$, $b = 6Cn/r$ and $c=2Cn$.
\end{proof}

\begin{theorem}
If each $c_t$ is $L$-Lipschitz, then for $n$ sufficiently large
and $\nu =\frac{R}{C\sqrt{n}}$, $\alpha = \frac{\delta}{r}$, and
$\delta = n^{-.25}\sqrt{\frac{RdCr}{3(Lr+C)}},$
     $$
     \E\bigg[\sum_{t=1}^n c_t(x_t)\bigg]
     -\min_{x \in S} \sum_{t=1}^n c_t(x)
     \leq 2n^{3/4}\sqrt{3RdC(L+C/r)}.$$
\end{theorem}
\begin{proof}
The proof is quite similar to the proof of Theorem
\ref{thm:param}. Again we check that the points $x_t \in S$, which
it is for $n$ is sufficiently large.  We now have a direct
Lipschitz constant, so we can use it directly in Eq.
(\ref{brendanwillneverreadthis}). Plugging this in with chosen
values of $\alpha$ and $\delta$ gives the lemma.
\end{proof}

\subsection{Reshaping}
The above regret bound depends on $R/r$, which can be very large.
  To
remove this dependence (or at least the dependence on $1/r$), we
can reshape the body to make it more ``round.''

The set $S$, with $r\B \subseteq S \subseteq R\B$ can be put in
{\em isotropic position} \cite{MP}. Essentially, this amounts to
estimating the covariance of random samples from the body and
applying an affine transformation $T$ so that the new covariance
matrix is the identity matrix.

A body $T(S) \subseteq \reals^d$ in isotropic position has several
nice properties, including $\B \subseteq T(S) \subseteq d\B$.  So,
we first apply the preprocessing step to find $T$ which puts the
body in isotropic position.  This gives us a new $R'=d$ and $r'=
1$.   The following observation shows that we can use $L'=LR$.
\begin{observation}
Let $c_t'(u) = c_t(T^{-1}(u))$.  Then $c_t'$ is $LR$-Lipschitz.
\end{observation}
\begin{proof}
Let $x_1, x_2 \in S$ and $u_1 = T(x_1)$, $u_2 = T(x_2)$.
Observe that,
$$|c_t'(u_1)-c_t'(u_2)| = |c_t(x_1)-c_t(x_2)| \leq
L\|x_1-x_2\|.$$
To make this a $LR$-Lipschitz condition on $c'_t$, it suffices to
show that
$\|x_1-x_2\| \leq R\|u_1-u_2\|.$ Suppose not, i.e.~$\|x_1 - x_2 \| > R\|u_1 - u_2\|$.
Define $v_1 = \frac{u_1 - u_2}{\| u_1 - u_2\|}$ and $v_2 = - v_1$.  Observe that $\|v_2 - v_1 \| = 2$, and since $T(S)$ contains the ball of radius $1$, $v_1, v_2 \in T(S)$.  Thus, $y_1 = T^{-1}(v_1)$ and $y_2 = T^{-1}(v_2)$ are in $S$.
Then, since $T$ is affine,
\begin{align*}
\| y_1 - y_2 \| = & \frac{1}{\|u_1 - u_2\|} \| T^{-1}(u_1 - u_2) - T^{-1}(u_2 - u_1)\| \\
 = & \frac{2}{\|u_1 - u_2\|} \| T^{-1}(u_1) - T^{-1}(u_2)\|\\
 = & \frac{2}{\|u_1 - u_2\|} \| x_1 - x_2\| > 2R,
\end{align*}
where the last line uses the assumption $\|x_1 - x_2 \| > R\|u_1 - u_2\|$.  The inequality $\|y_1 - y_2\| > 2R$ contradicts the assumption that $S$ is contained in a sphere of radius $R$.
\end{proof}

Many common shapes such as balls, cubes, etc.,
are
already nicely shaped, but there exist MCMC algorithms for putting
any body into isotropic position from a membership oracle
\cite{KLS,LV}. (Note that the projection oracle we assume is a
stronger oracle than a membership oracle.)  The latest (and
greatest) algorithm for putting a body into isotropic position,
due to Lovasz and Vempala \cite{LV}, runs in time $O(d^4)
\mbox{poly-log}(d,\frac{R}{r})$. This algorithm puts the body into
nearly isotropic position, which means that $\B \subseteq T(S)
\subseteq 1.01 d\B$.  After such preprocessing we would have
$r'=1,R'=1.01d,L'=LR,$ and $C'=C$. This gives,
\begin{corollary}
For a set $S$ of diameter $D$, and $c_t$ $L$-Lipschitz, after
putting $S$ into near- isotropic position, the BGD algorithm has
expected regret,
     $$\E\bigg[\sum_{t=1}^n c_t(x_t)\bigg]
     -\min_{x \in S} \sum_{t=1}^n c_t(x)
    \leq 6n^{3/4}d\left(\sqrt{CLR}+C\right).
    $$
Without the $L$-Lipschitz condition,
 $$
     \E\bigg[\sum_{t=1}^n c_t(x_t)\bigg]
     -\min_{x \in S} \sum_{t=1}^n c_t(x)
     \leq 6n^{5/6}dC$$
\end{corollary}
\begin{proof}
Using $r'=1,R'=1.01d,L'=LR,$ and $C'=C$, In the first case, we get
an expected regret of at most $ 2n^{3/4}\sqrt{6(1.01d)dC(LR+C)}$.
In the second case, we get an expected regret of at most  $3C
n^{5/6}\sqrt{2(1.01d)d}$.
\end{proof}

\subsection{Conclusions}
We have given algorithms for bandit online optimization of convex
functions.  Our approach is to extend Zinkevich's gradient descent
analysis to a situation where we do not have access to the
gradient.  We give a simple trick for approximating the gradient
of a function by a single sample, and we give a simple
understanding of this approximation as being the gradient of a
smoothed function.  This is similar to a similar approximation
proposed in \cite{S}. The simplicity of our approximation make it
straightforward to analyze this algorithm in an online setting,
with few assumptions.

Zinkevich presents a few nice variations on the model and
algorithms.  He shows that an adaptive step size $\eta_t =
O(1/\sqrt{t})$ can be used with similar guarantees.  It is likely
that a similar adaptive step size could be used here.

He also proves that gradient descent can be compared, to an
extent, with a non-stationary adversary.  He shows that relative
to any sequence $z_1,z_2,\ldots,z_n$, it achieves,
$$E\bigg[\sum_{t=1}^n c_t(x_t)\bigg]-\min_{z_1,z_2,\ldots,z_n \in S} \sum
c_t(z_t) \leq O\left(GD\sqrt{n(1+\sum \|z_t-z_{t-1}\|)}\right).$$
Thus, compared to an adversary that moves a total distance $o(n)$,
he has regret $o(n)$. These types of guarantees may be extended to
the bandit setting.

It would also be interesting to analyze the algorithm in an
unconstrained setting, where issues of the shape of the convex set
wouldn't come into play.  The difficulty is that in the
unconstrained setting we cannot assume the convex functions are
bounded. However, since $E[c_t(x_t+\delta
u_t)u_t]=E[\bigl(c_t(x_t+\delta u_t)-c_{t-1}(x_{t-1}+\delta u_{t-
1})\bigr)u_t]$,
 if the functions do not change too much from period
to period, one may be able to use the evaluation of the previous
period as a baseline to prevent the random gradient estimate from
being too large.

    \noindent
    {\bf Acknowledgements.}
We would like to thank David McAllester and Rakesh Vohra for
helpful discussions.  We are particularly grateful to Rakesh Vohra
for pointing us to the work of James Spall.

\end{document}